\documentclass[letterpaper, 10 pt, conference]{ieeeconf}  

\IEEEoverridecommandlockouts                              

\usepackage{graphics} 
\usepackage{epsfig} 
\usepackage{mathptmx} 
\usepackage{times} 
\usepackage{amsmath} 
\usepackage{amssymb}  
\usepackage[normalem]{ulem}
\usepackage[table]{xcolor}
\usepackage{booktabs}
\usepackage{multirow} 
\usepackage{pifont}
\usepackage{caption}
\usepackage{pifont}
\usepackage{bbm}
\usepackage{hyperref}

\title{\LARGE \bf
Talk2Escape: Conversational Grounding for Vision-and-Language Navigation
}

\author{Zerui Li$^{1}$$^{\dagger}$, Sihao Lin$^{1}$$^{\dagger\ddagger}$, Yanyan Shao$^{2}$, Jiwen Zhang$^{3}$, Xiangyu Shi$^{1}$, Shijie Li$^{4}$, Qi Wu$^{1\ddagger}$*
\thanks{\hangindent=1em $^{\dagger}$ Joint First Author \space\space\space\space\space\space\space\space\space\space\space\space* Corresponding Authors}%
\thanks{\hangindent=1em $^{1}$ Australian Institute for Machine Learning, Adelaide University}%
\thanks{\hangindent=1em $^{2}$ Zhejiang Wanli University}%
\thanks{\hangindent=1em $^{3}$ Fudan University}%
\thanks{\hangindent=1em $^{4}$ Agency for Science, Technology and Research (A*STAR), Shijie Li is supported by the Agency for Science, Technology and Research (A*STAR) under its Career Development Fund (Project No. H26-KSR0066)}%
\thanks{\hangindent=1em $^{\ddagger}$ Also with the Responsible AI Research Centre, Australian Institute for Machine Learning, Adelaide University}%
\thanks{\hangindent=1em Project page: \url{https://zeruili22.github.io/talk2escape/}}%
}

\begin{document}

\maketitle
\thispagestyle{empty}
\pagestyle{empty}

\begin{abstract}
While Vision-and-Language Navigation (VLN) has demonstrated remarkable success, the prevailing single-turn paradigm exposes a fundamental vulnerability: agents operate in a strictly open-loop manner. In practice, factors such as perceptual aliasing, sensor noise, and odometry drift can cause minor deviations to accumulate over time, often leading to catastrophic mission failures with no built-in mechanism for error recovery. To address this, we introduce \textit{Talk2Escape}, a proactive and model-agnostic dialogue intervention framework that reframes navigation as a closed-loop interactive process. At its core, a lightweight vision-language module continuously monitors agent kinematics. Upon detecting localized looping or severe trajectory divergence, it translates raw egocentric observations into concise, grounded queries to solicit targeted corrective feedback from either an algorithmic oracle or a human-in-the-loop. Extensive evaluations in high-fidelity simulators, including R2R-CE, RxR-CE, and VLNVerse,
 demonstrate that \textit{Talk2Escape} exhibits consistent improvements across diverse base agents. Empirically, \textit{Talk2Escape} achieves a 66.0\% Success Rate on R2R-CE, outperforming the current supervised and zero-shot state-of-the-art methods. We further validate its sim-to-real transfer on a Unitree Go2 quadruped, proving that proactive dialogue drastically improves navigation robustness in physical environments.
\end{abstract}
 
\section{Introduction}
\label{sec:intro}

\begin{figure}[t]
  \centering
  \includegraphics[width=0.95\linewidth]{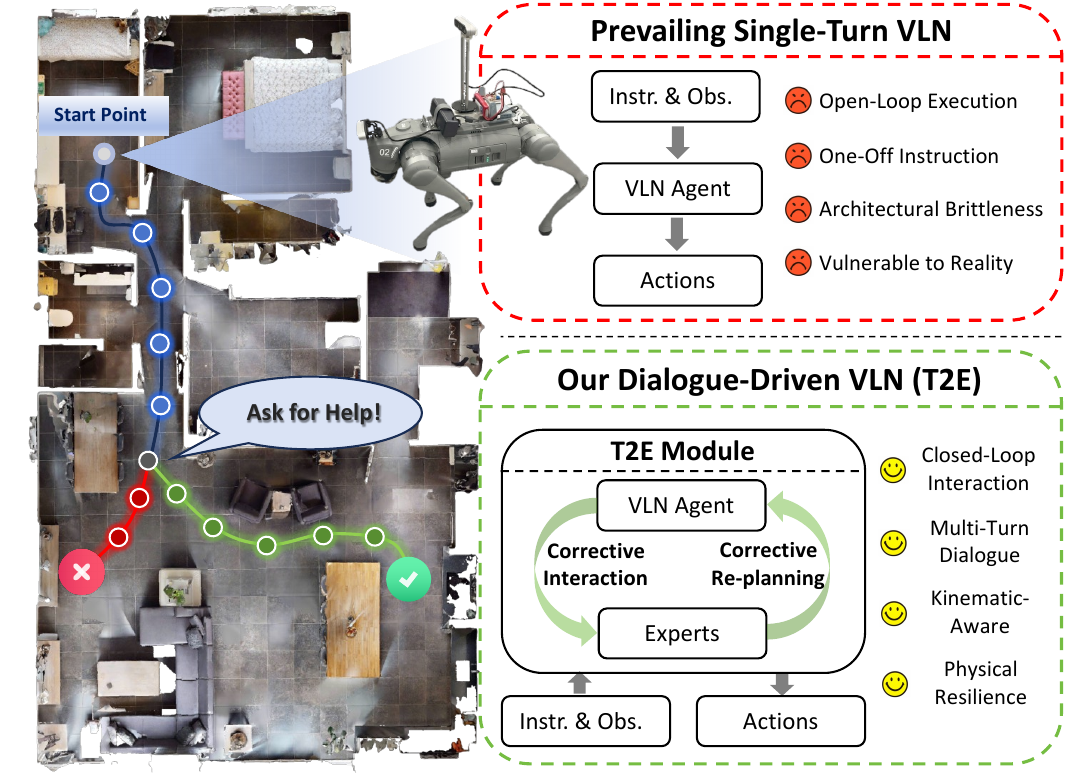}
  \vspace{-5pt}
  \caption{\textbf{Paradigm shift from single-turn to failure-aware dialogue-driven VLN:} \textbf{Left:} Single-turn agents operate in an open-loop manner, making them architecturally brittle to physical uncertainties. \textbf{Right:} Our \textit{Talk2Escape} framework introduces a closed-loop, proactive error-recovery mechanism, which actively solicits corrective feedback from a human guide, effectively rescuing the agent from severe real-world execution failures.} 
  \label{fig:teaser}
  \vspace{-2em}
\end{figure}

Vision-and-Language Navigation (VLN)~\cite{anderson2018vision} has demonstrated remarkable success rates within high-fidelity simulators, driven by the rapid advancement of Multimodal Large Language Models (MLLMs)~\cite{zhou2024navgpt, chen2024mapgpt, qiao2025open, zhang2025embodied}. However, directly deploying these learned policies onto physical robots reveals a  generalization gap. As shown in Fig.~\ref{fig:teaser}, the cause lies in the prevailing single-turn paradigm: agents operate in an open-loop manner, executing a one-off instruction without any mechanism to verify progress, seek clarification, or recover from errors~\cite{hong2022bridging, an2024etpnav, zhang2024navid, cheng2024navila}. This architectural design becomes particularly brittle under real-world conditions, where sensor noise, odometry drift, and unmapped obstacles are unavoidable.

In real-world settings, where sensor noise, odometry drift, and unmapped obstacles are unavoidable, even minor locomotion slips can push the robot into an unrecoverable ``lost'' state, leading to task failure. Beyond the inherent limitations of perception or planning, this catastrophic brittleness is largely architectural, as an open-loop system lacks the proactive mechanism to recover from inevitable physical errors.

How do humans maintain robustness during navigation? When individuals get lost in unfamiliar buildings, they rarely continue wandering aimlessly until completely failing their objectives. Instead, they instinctively initiate an active help-seeking process by describing their current visual surroundings to a guide and requesting corrective instructions. This intuitive error-recovery strategy is absent in standard robotic navigation frameworks. Recently, dialogue-driven VLN has been introduced in continuous environment simulators to facilitate multi-turn interactions. However, existing works have primarily utilized dialogue as supplementary context to refine initial planning in simulated environments, leaving its potential as a proactive error-recovery mechanism in physical deployments largely underexplored~\cite{long2024discuss, lin2025vlnverse, qiao2023march, li2024human, han2025dialnav, dong2025ha}. Compounding this limitation, these simulated benchmarks often abstract away physical execution errors and assume instant, perfect communication~\cite{wang2025rethinking, lin2025vlnverse}. 

We argue that dialogue interaction is valuable as an operational error-recovery channel that supports robust navigation under real-world uncertainties~\cite{lin2025vlnverse}. Motivated by this paradigm shift, we introduce \textit{Talk2Escape (T2E)}, a deployable, failure-aware framework that leverages the profound zero-shot reasoning capabilities of off-the-shelf MLLMs. Operating as a strictly model-agnostic module, our approach represents the first attempt to explore the efficacy of pure zero-shot MLLMs for interactive, dialogue-driven navigation on physical robots. It equips existing open-loop agents with an active error-recovery mechanism ~\cite{long2024discuss, liu2025nav}. A dedicated vision-language translator continuously monitors the agent's spatial kinematics. Upon detecting localized looping or severe trajectory divergence, it grounds rich egocentric visual observations into concise, navigation-relevant natural language queries. Through this mechanism, the robot proactively pauses and solicits targeted corrective feedback from an algorithmic oracle or a human-in-the-loop. By treating dialogue not as an evaluation metric, but as a critical operational safety net, \textit{Talk2Escape} enables proactive, multi-turn error recovery for unmapped real-world deployment.

Our main contributions are summarized as follows:

\begin{itemize}
    \item We propose \textit{T2E}, the first attempt to explore the deployment of zero-shot MLLMs for dialogue-driven VLN on physical robots, re-framing multi-turn interaction as a proactive error-recovery mechanism. 
    \item We formulate a condition-driven mechanism that bridges physical anomalies with conversational reasoning, dynamically translating spatial deadlocks into grounded queries.
    \item We empirically demonstrate that \textit{T2E} rescues agents from severe real-world failures, identifying the critical ``asymmetric context'' bottleneck for future embodied AI.
\end{itemize}

\section{Related Work}

\begin{figure*}[t]
  \centering
  \vspace{1em}
  \setlength{\abovecaptionskip}{2pt}
  \setlength{\belowcaptionskip}{3pt}
  \includegraphics[width=0.95\linewidth]{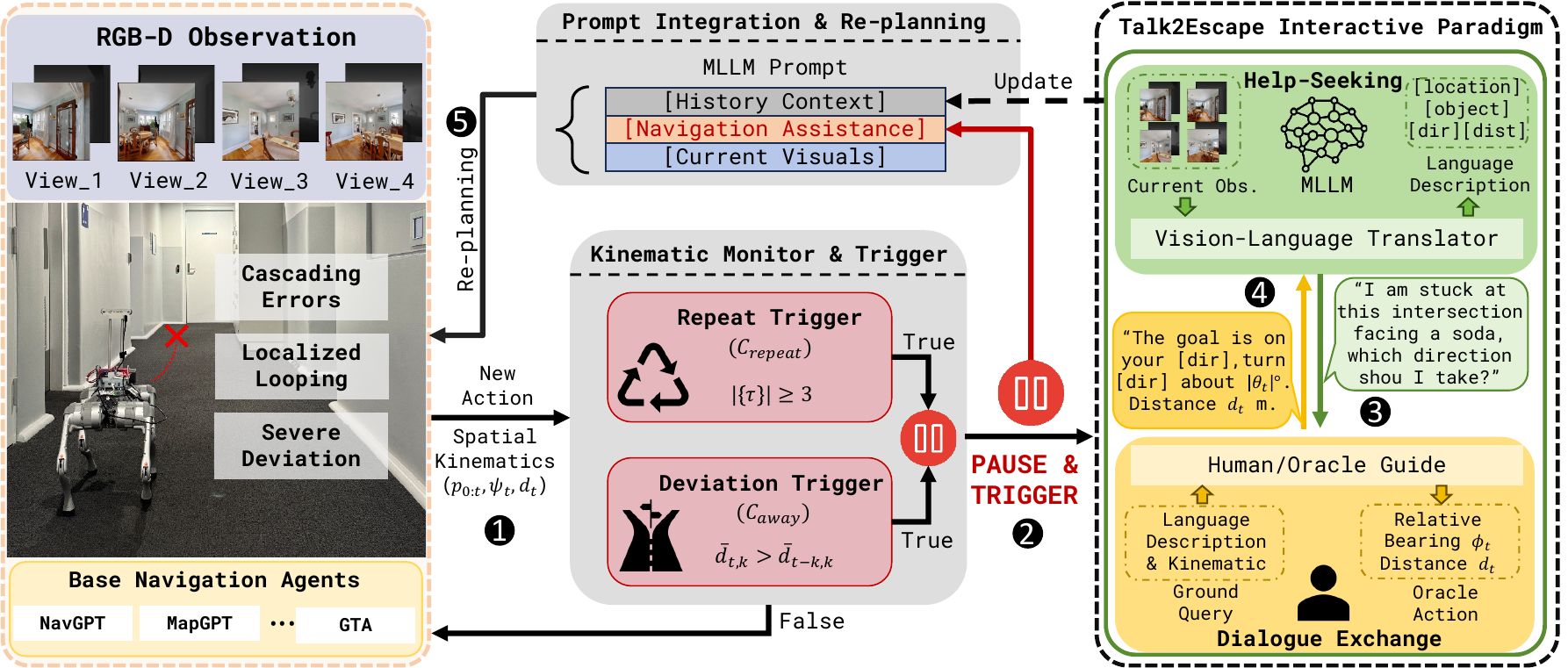}
  \caption{\textbf{Overview of the \textit{Talk2Escape} Framework:} \ding{182} Base navigation agents output spatial kinematics and action during execution. \ding{183} The Kinematic Monitor continuously evaluates these states. If cascading errors occur, it activates a \textit{Repeat} ($\mathcal{C}_{\mathrm{rep}}$) or \textit{Deviation} ($\mathcal{C}_{\mathrm{away}}$) trigger, proactively pausing the robot and sending the error state copy to the \textit{Navigation Assistance} block. \ding{184} The Vision-Language Translator grounds current RGB-D observations into a concise, active query. \ding{185} A Oracle Guide responds with a targeted corrective hint based on relative bearing and distance. \ding{186} This entire dialogue exchange is injected into the MLLM prompt, empowering the agent to re-plan its actions and successfully escape the unrecoverable state.}

  \label{fig:main}
  \vspace{-1em}
\end{figure*}

\subsection{Vision-and-Language Navigation (VLN)}
As a cornerstone of Embodied AI, Vision-and-Language Navigation (VLN) requires agents to interpret visual inputs and execute natural language instructions to navigate complex environments. Early research predominantly relied on discrete benchmarks~\cite{anderson2018vision, fried2018speaker, qi2020reverie}, abstracting physical movement into simplistic node-to-node "teleportation" along predefined topological graphs. To bridge the gap toward practical robotics, the field shifted to Continuous Environments (VLN-CE)~\cite{krantz2020beyond, irshad_hierarchical_2021, irshad_sasra_2021, raychaudhuri_language-aligned_2021}, replacing these topological jumps with unconstrained, low-level kinematic control. While executing navigation policies in continuous simulators already poses significant challenges, directly deploying these standard single-turn frameworks to real-world settings exposes a more fundamental bottleneck: the inherent inability to recover from inevitable physical execution errors. To address this open-loop vulnerability, researchers have begun to look beyond one-off instructions, motivating the emergence of dialogue-driven VLN.

\subsection{Zero-Shot Embodied Agents in VLN}
To mitigate the decision-making complexities inherent in continuous spaces, earlier research primarily relied on task-specific expert models and waypoint-based architectures~\cite{hong2022bridging, krantz2021waypoint, li2025ground, wang2023gridmm, an2024etpnav, an2023bevbert}. While these methods improved benchmark success rates, their policies are tightly coupled to specific training distributions, limiting their generalization to unseen environments. To overcome this, recent works have introduced Zero-Shot Embodied Agents powered by Multimodal Large Language Models (MLLMs)~\cite{qiao2025open, shi2025smartway, shi2025fast, li2025boosting, li2026one}. By harnessing the broad, open-vocabulary reasoning capabilities of foundation models, these agents can interpret instructions and navigate environments without domain-specific training. However, existing zero-shot VLN frameworks are strictly constrained to a single-turn paradigm. They execute initial instructions in an open-loop manner, completely lacking mechanisms to verify progress or recover from the physical disruptions inevitable in real-world deployment. Consequently, there is a critical need to unlock the intrinsic dialogue capabilities of zero-shot MLLMs, transitioning them from rigid single-turn executors into interactive agents capable of multi-turn conversational error recovery.

\subsection{Dialogue-Driven VLN}
To mitigate the brittleness of single-turn instruction following, dialogue-driven VLN was introduced to enable multi-turn human-agent interactions~\cite{qiao2023march, long2024discuss}. While recent advancements have extended these interactive frameworks into continuous simulators~\cite{long2024discuss, lin2025vlnverse, li2024human, han2025dialnav, dong2025ha}, existing approaches overwhelmingly rely on supervised learning, reinforcement learning, or parameter-efficient fine-tuning on specific conversational datasets. They inherently treat dialogue as a learned feature mapping rather than a generalized cognitive process. Consequently, these models are deeply anchored to simulated environments, where communication is assumed instant and physical execution errors are non-existent, making their transition to real-world robotics highly challenging. In contrast, \textit{Talk2Escape} bypasses conversational fine-tuning and instead leverages off-the-shelf MLLMs as a zero-shot reasoning engine for physical error recovery. By formulating multi-turn interaction as an on-the-fly, zero-shot reasoning process, we empower real-world robots to actively generate context-aware queries and survive unmapped physical disruptions, moving beyond the constraints of pre-trained dialogue templates.
 
\section{Preliminary}
\label{sec:preliminary}

\subsection{Problem Formulation}
\label{subsec:problem_formulation}

We first formalize the standard single-turn Vision-and-Language Navigation (VLN) task in continuous, unmapped environments. The agent is initialized at a starting pose $p_0 = (x_0, y_0, \theta_0)$ and is provided with a natural language instruction $\mathcal{I}$. At each discrete reasoning step $t$, the agent obtains its current estimated pose $p_t$ and a structured egocentric observation $O_t$. 

To ensure comprehensive environmental awareness for the reasoning module, the observation $O_t$ consists of four orthogonal RGB-D views captured at cardinal angles relative to the current heading of the agent:
$$O_t = \left\{ v_t^k \mid k \in \{0^\circ, 90^\circ, 180^\circ, 270^\circ\} \right\}$$
This panoramic configuration offers a $360^\circ$ visual context with minimal redundancy. The agent maintains a state history $H_t = \{\mathcal{I}, (O_1, p_1, a_1), \dots, (O_t, p_t)\}$, encapsulating the initial instruction and the sequence of past experiences. 

The objective is to derive a high-level policy $\pi$ that maps the history $H_t$ to a physical action $a_t \in \mathcal{A}^p$. To accommodate diverse base architectures (e.g., NavGPT, GTA), the physical action space $\mathcal{A}^p$ can encompass structured discrete action tokens, continuous relative waypoints $(\Delta x, \Delta y)$, or normalized pixel coordinates $(u, v)$. A low-level controller then executes a trajectory to reach this target. However, in this prevailing single-turn paradigm, the instruction $\mathcal{I}$ remains strictly static. When the agent encounters navigation uncertainties, perceptual aliasing, or physical execution errors, it operates in a vulnerable open-loop manner, lacking any external mechanisms to recover from the resulting spatial discrepancies.

\subsection{Dialogue-Driven Navigation}
\label{subsec:dialogue_navigation}

To mitigate the architectural brittleness of open-loop execution, Dialogue-Driven Navigation reframes the task as an interactive, multi-turn process. Instead of being restricted solely to physical movements, the agent's capabilities are augmented with a communication action space $\mathcal{A}^c$. 

During navigation, if the agent encounters severe uncertainties or detects predefined anomalous behaviors, it can suspend its physical movement and execute a communication action $a^c \in \mathcal{A}^c$. This initiates a Help-Seeking phase, triggering the generation of a grounded natural language query $Q_k$. Derived from the comprehensive observation $O_t$, this query articulates the current spatial confusion. The query is then transmitted to an external guide or Oracle $\mathcal{G}$, who provides targeted corrective feedback $F_k$. 

Consequently, the navigation process is no longer conditioned on a one-off static instruction, but rather on an evolving, multi-turn dialogue context. Let $D_k = \{(Q_1, F_1), \dots, (Q_k, F_k)\}$ denote the dialogue history up to the $k$-th interaction. The updated navigation history becomes $H'_t = H_t \cup D_k$, and the agent policy is redefined as a robust closed-loop system:
\vspace{-1em}
$$a_t = \pi(H'_t)$$\vspace{-2em}

In our proposed \textit{Talk2Escape} framework, rather than learning $\pi$ through rigid and resource-intensive parameter fine-tuning, we instantiate this dynamic multi-turn dialogue policy directly by harnessing the zero-shot multimodal reasoning capabilities of frozen Multimodal Large Language Models (MLLMs). 
\section{Methodology}
\label{sec:method}

\subsection{Base Navigation Agent Configuration}
\label{subsec:navigation_framework}

Our proposed \textit{Talk2Escape} intervention mechanism is inherently model-agnostic. To contextualize its operation, we describe its instantiation upon a comprehensive continuous navigation agent, such as NavGPT~\cite{zhou2024navgpt}, MapGPT~\cite{chen2024mapgpt} and  GTA~\cite{li2026one}. At each reasoning step $t$, the base agent acquires egocentric RGB-D observations. The Multimodal Large Language Model (MLLM) processes these visuals and specifies navigation targets via normalized pixel coordinates $(u, v)$.

To enable long-horizon spatial awareness, advanced agents typically supplemented by an incrementally constructed TSDF top-down map and maintain a topological graph $\mathcal{G} = (\mathcal{V}, \mathcal{E})$ that tracks visited locations. While such spatial reasoning modules provide foundational capabilities, they remain strictly open-loop; they cannot recover when the MLLM's spatial logic fatally collapses. This fundamental limitation necessitates an external, proactive dialogue intervention.

\subsection{Talk2Escape: Proactive Dialogue Intervention}
\label{sec:dialogue}

A critical vulnerability of zero-shot MLLM navigation is the inability to recover from cascading errors. Once an agent drifts off course or enters a perceptual loop, the lack of external corrective signals inevitably leads to mission failure. We propose the \textbf{Talk2Escape} mechanism as shown in Fig.~\ref{fig:main}: an interactive paradigm that monitors kinematics, translates failures into grounded questions, and injects targeted navigational hints, effectively simulating an over-the-shoulder human guide.

\subsubsection{Intervention Trigger Conditions}
We define six intervention strategies to systematically study the timing and necessity of guidance:

\begin{itemize}
    \item \textbf{None} (Baseline): No intervention is provided, establishing the strictly unassisted zero-shot lower bound.
    
    \item \textbf{Repeat Trigger} ($\mathcal{C}_{\mathrm{rep}}$): Activated when the agent enters a localized loop. It is triggered if the agent visits the current spatial region (radius $r_{\mathrm{rep}} = 0.5\,\text{m}$) at least $n_{\mathrm{rep}} = 3$ times. We elegantly define this using set cardinality:
    \begin{equation}
            \mathcal{C}_{\mathrm{rep}}(t) = \mathbb{1} \Big[ \big| \{ \tau \in [0, t] \mid \|p_\tau - p_t\|_{xz} < r \} \big| \geq n \Big]
    \end{equation}
    where $\mathbb{1}[\cdot]$ denotes the indicator function that evaluates to 1 if the enclosed condition is true and 0 otherwise, $\|\cdot\|_{xz}$ represents the Euclidean distance on the horizontal plane, $r = 0.5\,\text{m}$ is the radius of the spatial region, and $n = 3$ is the visit count threshold.
    
    \item \textbf{Deviation Trigger} ($\mathcal{C}_{\mathrm{away}}$): Activated when the agent actively diverges from the target. To mitigate noise from single-step kinematic jitter, we compare the moving average distance to the goal over a sliding window of $k = 2$ steps. Let $d_t = \|p_t - p_g\|_{xz}$. The trigger is defined as:
    \begin{equation}
        \begin{aligned}
        \mathcal{C}_{\mathrm{away}}(t) = \mathbb{1} \Big[ \bar{d}_{t, k} > \bar{d}_{t-k, k} \Big], \quad\\
        \text{where } \bar{d}_{t, k} = \frac{1}{k}\sum_{i=0}^{k-1} d_{t-i}
        \end{aligned}
    \end{equation}
    
    \item \textbf{Hybrid Trigger} (\textbf{Both}): Defined as $\mathcal{C}_{\mathrm{both}}(t) = \mathcal{C}_{\mathrm{rep}}(t) \lor \mathcal{C}_{\mathrm{away}}(t)$. This provides a comprehensive safety net against both looping and divergence.
    
    \item \textbf{Periodic}: A hint is issued strictly every $k_{\mathrm{per}} = 5$ steps, testing if rigid scheduled guidance is effective.
    
    \item \textbf{Human-in-the-Loop (Human)}: A human supervisor is synchronously polled using the agent's discrete visuals to issue interventions, establishing a practical subjective baseline.
\end{itemize}

\subsubsection{Vision-Language Translator for Help-Seeking}
When an automated trigger ($\mathcal{C}_{\mathrm{rep}}$ or $\mathcal{C}_{\mathrm{away}}$) is activated, the system does not merely force a hint onto the agent. Instead, it must transform raw multi-view perception into a concise, active query that a human/oracle guide can answer reliably. 

We employ an MLLM-based translator that: (i) summarizes the local scene from the current RGB-D views, (ii) identifies the most plausible failure mode given the trigger condition (e.g., looping in a corridor, facing a dead end, or stalling at an ambiguous intersection), and (iii) produces a short, grounded question requesting a directional correction (e.g., \textit{``I am stuck at this intersection facing a sofa; which direction should I take?''}). The translator is strictly constrained to reference observable landmarks and relative directions, avoiding speculative statements about unseen topological areas.

\subsubsection{Oracle Hint Generation and Integration}
In response to the agent's grounded query, the Oracle system computes the optimal corrective hint. Let $\psi_t$ denote the agent's current heading and $\phi_t = \mathrm{atan2}(p_g^z - p_t^z,\; p_g^x - p_t^x)$ the absolute bearing to the goal. The relative turn angle is $\theta_t = \phi_t - \psi_t \in (-180^{\circ}, 180^{\circ}]$. The generated response $h_t$ is formulated as:
\vspace{-4pt}
\begin{equation}
    h_t =
    \begin{cases}
        \begin{array}{@{}l@{}}
            \text{``You are very close to} \\
            \text{the goal.''}
        \end{array} & \text{if } d_t < d_{\mathrm{th}} \\[4ex]
        \begin{array}{@{}l@{}}
            \text{``The goal is on your [dir],} \\
            \text{turn [dir] about $|\theta_t|^{\circ}$.} \\
            \text{Distance to goal: $d_t$\,m.''}
        \end{array} & \text{otherwise}
    \end{cases}
\end{equation}
where $[\text{dir}] \in \{\text{left}, \text{right}\}$ is determined by the sign of $\theta_t$. 

Finally, this dialogue exchange—comprising the agent's proactive query and the oracle's hint $h_t$—is injected into the MLLM prompt as a dedicated \textit{Navigation Assistance} block. Placed strategically between the historical context and the current visual inputs, this dialogue provides the agent with an unambiguous corrective directive to escape its current failure mode. 
\section{Experiments}
\label{sec:experiments}
\begin{table*}[t]
\centering
\setlength{\abovecaptionskip}{2pt}
\setlength{\belowcaptionskip}{3pt}
\small
\captionsetup{skip=5pt}
\setlength{\tabcolsep}{9pt}

\caption{\textbf{Comparison in continuous environments on R2R-CE and RxR-CE Val-Unseen splits.} 
The \textbf{best supervised} results are highlighted in \textbf{bold}, while the \underline{best zero-shot} results are \underline{underlined}. 
We report metrics on sampled subsets for R2R-CE (100 episodes, following Open-Nav~\cite{qiao2025open}) and RxR-CE (260 episodes), following GTA~\cite{li2026one}.}
\label{tab:continuous_main}

\resizebox{0.9\linewidth}{!}{
\begin{tabular}{cl ccccc cccc}
\toprule

\multirow{2}{*}{\textbf{\#}} & \multirow{2}{*}{\textbf{Methods}} & \multicolumn{5}{c}{\textbf{R2R-CE}} & \multicolumn{4}{c}{\textbf{RxR-CE}} \\
\cmidrule(lr){3-7} \cmidrule(lr){8-11} 
& & \textbf{NE}$\downarrow$ & \textbf{OSR}$\uparrow$ & \textbf{SR}$\uparrow$ & \textbf{SPL}$\uparrow$ & \textbf{nDTW}$\uparrow$ & \textbf{NE}$\downarrow$ & \textbf{SR}$\uparrow$ & \textbf{SPL}$\uparrow$ & \textbf{nDTW}$\uparrow$ \\
\midrule
\rowcolor{blue!5} \multicolumn{11}{l}{\textbf{\textit{Supervised Learning:}}} \\

1 & CMA~\cite{hong2021vln} & 6.30 & 49.0 & 38.0 & 33.0 & -- & 10.4 & 24.1 & 19.1 & 37.4 \\
2 & VLN-BERT~\cite{hong2021vln} & 5.74 & 53.0 & 44.0 & 39.0 & -- & 8.98 & 27.1 & 22.7 & 46.7 \\
3 & GridMM~\cite{wang2023gridmm} & 5.11 & 61.0 & 49.0 & 41.0 & -- & -- & -- & -- & -- \\
4 & ETPNav~\cite{an2024etpnav} & 4.71 & 65.0 & 57.0 & 49.0 & -- & 5.64 & 54.8 & 44.9 & 61.9 \\
5 & BEVBert~\cite{an2023bevbert} & 4.70 & 67.0 & 59.0 & 50.0 & -- & 4.80 & 64.4 & -- & 65.4 \\
6 & HNR~\cite{wang2024lookahead} & 4.42 & 67.0 & 61.0 & 51.0 & -- & 5.51 & 56.4 & 46.7 & 63.6 \\
7 & NavFoM~\cite{zhang2025embodied} & 4.61 & 72.1 & 61.7 & 55.3 & -- & 4.74 & 64.4 & \textbf{56.2} & 65.8 \\
8 & Efficient-VLN~\cite{zheng2025efficient} & \textbf{4.18} & \textbf{73.7} & \textbf{64.2} & \textbf{55.9} & -- & \textbf{3.88} & \textbf{67.0} & 54.3 & \textbf{68.4} \\
\midrule
\rowcolor{blue!5} \multicolumn{11}{l}{\textbf{\textit{Zero-Shot Learning:}}} \\
9 & Open-Nav~\cite{qiao2025open} & 6.70 & 23.0 & 19.0 & 16.1 & 45.8 & -- & -- & -- & -- \\
10 & CA-Nav~\cite{chen2025constraint} & 7.58 & 48.0 & 25.3 & 10.8 & -- & 10.37 & 19.0 & 6.0 & -- \\
11 & SmartWay~\cite{shi2025smartway} & 7.01 & 51.0 & 29.0 & 22.5 & -- & -- & -- & -- & -- \\
12 & STRIDER~\cite{he2025strider} & 6.91 & 39.0 & 35.0 & 30.3 & 51.8 & 11.19 & 21.2 & 9.6 & 30.1 \\
13 & VLN-Zero~\cite{bhatt2025vln} & 5.97 & 51.6 & 42.4 & 26.3 & -- & 9.13 & 30.8 & 19.0 & -- \\
14 & BZS-VLN~\cite{li2025boosting} & 6.12 & 55.0 & 41.0 & 25.4 & -- & 7.56 & 35.7 & 21.7 & 42.4 \\
15 & GTA~\cite{li2026one} & 4.95 & 56.2 & 48.8 & 41.8 & \underline{60.4} & 6.29 & 46.2 & \underline{39.3} & \underline{57.4} \\

\midrule 
\rowcolor{gray!10}
16 & \textbf{Talk2Escape + NavGPT Agent}
& 4.84 & 60.0 & 64.0 & 46.4 & 49.9
& 6.01 & 50.4 & 27.2 & 45.7 \\
\rowcolor{gray!10}
17 & \textbf{Talk2Escape + GTA Agent}
& \underline{4.80} & \underline{66.0} & \underline{72.0} & \underline{49.4} & 51.4
& \underline{5.89} & \underline{62.9} & 34.2 & 49.9 \\

\bottomrule
\end{tabular}
}
\vspace{-4mm}
\end{table*}

\begin{table}[t]
\centering
\caption{\textbf{Zero-Shot Performance on VLNVerse.} Evaluation on the fine-grained instruction task, comparing standard open-loop agents against their passive Chain-of-Thought (CoT) variants and our proposed interactive \textit{T2E} module.}
\label{tab:vlnverse_abl}
\resizebox{\columnwidth}{!}{
\begin{tabular}{l ccccc} 
\toprule
\textbf{Prompting Strategy} & \textbf{NE ($\downarrow$)} & \textbf{SR ($\uparrow$)} & \textbf{OSR ($\uparrow$)} & \textbf{SPL ($\uparrow$)} & \textbf{nDTW ($\uparrow$)} \\
\midrule
NavGPT~\cite{zhou2024navgpt} & 6.10 & 19.30 & 61.50 & 8.13 & 52.60 \\
NavGPT~\cite{zhou2024navgpt} + CoT & 5.25 & 28.70 & 50.50 & 9.91 & 44.10 \\
NavGPT~\cite{zhou2024navgpt} + \textbf{T2E} & \textbf{2.92} & \textbf{84.60} & \textbf{88.30} & \textbf{35.20} & \textbf{60.50} \\
\midrule
MapGPT~\cite{chen2024mapgpt} & 5.62 & 25.53 & 55.32 & 7.57 & 43.38 \\
MapGPT~\cite{chen2024mapgpt} + CoT & 4.51 & 42.19 & 72.40 & 11.32 & 45.20 \\
MapGPT~\cite{chen2024mapgpt} + \textbf{T2E} & 4.14 & 67.02 & 74.47 & 30.46 & 53.62 \\
\bottomrule
\end{tabular}
}
\end{table}

\begin{table}[t]
    \centering
    \small
    \captionsetup{skip=5pt}
    \setlength{\tabcolsep}{8pt}
    
    \caption{\textbf{Real-world navigation performance.} We compare Talk2Escape against representative supervised (VLN-BERT, RDP) and zero-shot (SmartWay) baselines. $\uparrow$ indicates higher is better, $\downarrow$ indicates lower is better.}
    \label{tab:real_robot_results}
    
    \begin{tabular}{l cc}
        \toprule
        \textbf{Method} & \textbf{SR}$\uparrow$ (\%) & \textbf{NE}$\downarrow$ (m) \\
        \midrule
        
        \rowcolor{blue!5} \multicolumn{3}{l}{\textbf{\textit{Supervised Learning:}}} \\
        VLN-BERT~\cite{hong2021vln} & 16.0 & 5.36 \\
        RDP~\cite{wang2025rethinking} & 20.0 & 5.45 \\ 
        
        \midrule

        \rowcolor{blue!5} \multicolumn{3}{l}{\textbf{\textit{Zero-Shot Learning:}}} \\
        SmartWay~\cite{shi2025smartway} & 32.0 & 4.85 \\ 
        GTA & \textbf{40.0} & \textbf{3.66} \\
        \textbf{Talk2Escape + GTA Agent} & \textbf{62.0} & \textbf{3.31} \\
        \midrule
        \bottomrule
    \end{tabular}
    \vspace{-2em} 
\end{table}
To rigorously assess the proposed \textit{Talk2Escape} framework, we design comprehensive experiments to answer four core research questions: 
\begin{itemize}
    \item \textbf{RQ1:} How does our interactive zero-shot navigation framework compare against existing state-of-the-art supervised and zero-shot methods in complex continuous environments?
    
    \item \textbf{RQ2:} Is the proposed mechanism model-agnostic, and how effectively does it elevate the performance of structurally diverse base agents (from vanilla baselines to highly optimized SOTA frameworks)?
    
    \item \textbf{RQ3:} To what extent does the proactive help-seeking mechanism trade off optimal path efficiency (e.g., SPL, nDTW) against absolute navigation success (SR) in excessively long trajectories?
    
    \item \textbf{RQ4:} Can this method effectively to real-world physical quadruped robots coping with asynchronous control and dynamic sensor noise?
\end{itemize}

\subsection{Experimental Setup}
\label{subsec:setup}

\subsubsection{Datasets and Simulators}
We evaluate our method in the Matterport3D continuous environment simulator primarily using two standard benchmarks: \textbf{R2R-CE}~\cite{krantz2020beyond} and \textbf{RxR-CE~\cite{ku2020room}}. R2R-CE transfers the original Room-to-Room discrete paths into continuous trajectories. RxR-CE is significantly more challenging, featuring longer and more spatially complex instructions. To ensure a strictly fair and rigorous comparison, we adopt the exact evaluation protocol established by Open-Nav~\cite{qiao2025open}. Specifically, we report our primary metrics on a representative subset of 100 episodes sampled from the R2R-CE Validation-Unseen split. Additionally, for the RxR-CE benchmark, following GTA~\cite{li2026one}, we report performance on a sampled subset of 260 episodes.

\paragraph{Evaluation Metrics}
We report standard metrics for VLN-CE Task~\cite{krantz2020beyond}: \textit{Success Rate} (SR), \textit{Oracle Success Rate} (OSR), \textit{Success weighted by Path Length} (SPL), \textit{Trajectory Length} (TL), and \textit{Navigation Error} (NE). Additionally, we compute \textit{normalized Dynamic Time Warping} (nDTW) to assess the fidelity of the agent's trajectory relative to the ground truth path. An episode is considered successful if the agent stops within 3.0 meters of the target coordinates.

\subsubsection{Implementation Details}
Our framework leverages Gemini 3.1 Pro as the core multimodal reasoning engine. To demonstrate that the proposed \textit{Talk2Escape} mechanism is fundamentally model-agnostic, we instantiate it upon two representative zero-shot base agents: NavGPT~\cite{zhou2024navgpt} and GTA~\cite{li2026one}. NavGPT is adopted as a vanilla baseline; its fundamental, auxiliary-free architecture provides a rigorous testbed to strictly isolate the performance gains attributed solely to our interactive dialogue intervention. Conversely, GTA represents the current state-of-the-art (SOTA) in zero-shot continuous VLN, allowing us to evaluate whether proactive dialogue can further elevate the upper bound of highly optimized systems. For the intervention triggers, the proximity merge threshold $\delta$ is set to $0.5\,\text{m}$. The \textit{Repeat} trigger is activated using a radius $r_{\mathrm{rep}} = 0.5\,\text{m}$ over $n_{\mathrm{rep}} = 3$ steps, while the \textit{Deviation} trigger monitors trajectory divergence over a sliding window of $k_{\mathrm{away}} = 2$ steps.

\subsubsection{Real-World Robot Deployment}
To validate sim-to-real transferability, we deploy \textit{Talk2Escape} on a Unitree Go2 quadruped equipped with an active perception payload: a servo-mounted depth camera that autonomously sweeps to acquire orthogonal views at each reasoning step. To ensure smooth locomotion during MLLM inference, we explicitly decouple cognition from execution. The MLLM asynchronously computes local target waypoints at a lower frequency, dispatching them to the robot's native point-to-point navigation stack for high-frequency, continuous velocity tracking. Finally, to maintain experimental parity, we enforce a strict human-in-the-loop (HITL) protocol. A human expert is polled every cycle but is procedurally constrained to provide hints \textit{only} upon explicitly observing predefined failure modes, rigorously isolating the dialogue mechanism's efficacy from arbitrary teleoperation.

\subsection{Main Results: Comparison with State-of-the-Art}
\label{subsec:main_results}

To answer RQ1, we compare our complete \textit{Talk2Escape} framework against a comprehensive suite of prior methods, including heavily fine-tuned \textit{Supervised Learning} baselines and recent \textit{Zero-Shot Learning} architectures. Tab.~\ref{tab:continuous_main} presents the quantitative results on the R2R-CE and RxR-CE Val-Unseen splits. As shown in Tab.~\ref{tab:continuous_main}, integrating our proactive dialogue mechanism yields unprecedented navigation performance. Remarkably, \textit{Talk2Escape + GTA Agent} achieves an absolute state-of-the-art Zero-Shot Success Rate (SR) of 72.0\% on R2R-CE, outperforming the previous best open-loop zero-shot method (GTA at 48.8\%) by a massive margin of +23.2\%. More importantly, without observing a single in-domain training trajectory, our interactive framework surpasses even the highest-performing supervised SOTA, Efficient-VLN (64.2\% SR). This demonstrates that closed-loop, language-driven error recovery can fundamentally break the performance ceiling of traditional open-loop embodied agents.

The architectural universality of our framework is validated by the \textit{Talk2Escape + NavGPT Agent}. Despite employing a vanilla, unoptimized base agent, it achieves a 64.0\% SR on R2R-CE, surpassing advanced zero-shot models such as VLN-Zero (42.4\%) and GTA (48.8\%), and approaching the best supervised methods. This suggests that the capability to actively seek help provides a higher performance floor than passive architectural engineering alone. On the more challenging RxR-CE benchmark, we observe an expected trade-off. While \textit{Talk2Escape + GTA Agent} raises the SR from 46.2\% to 62.9\%, SPL decreases from 39.3\% to 34.2\% and nDTW from 57.4\% to 49.9\%. This is a direct consequence of the recovery process rather than degraded navigation quality. Rescuing a divergent agent inherently involves backtracking and exploratory detours, which lengthen the executed path and, by construction, penalize path-efficiency metrics. In other words, episodes that would otherwise fail outright are converted into successes at the cost of longer trajectories. We consider this a worthwhile trade-off, as it prioritizes mission completion over path optimality in long-horizon environments.

\subsection{Ablation Studies}
\label{subsec:ablation}

\begin{figure}[t] 
    \centering
   
    \begin{minipage}{\linewidth}
        \centering
        \setlength{\abovecaptionskip}{2pt}
        \setlength{\belowcaptionskip}{3pt}
        \small
        \captionsetup{type=table} 
        \caption{\textbf{Trigger Conditions vs. Performance:} Comparison of various dialogue trigger strategies, evaluations are conducted on the same 100 sampled episodes of R2R-CE as in Tab.~\ref{tab:continuous_main}.}
        \label{tab:ablation_triggers}
        \resizebox{\linewidth}{!}{
        \begin{tabular}{l ccc c}
        \toprule
        \textbf{Trigger Strategy} & \textbf{SR ($\uparrow$)} & \textbf{SPL ($\uparrow$)} & \textbf{Count} & \textbf{Rate} \\
        \midrule
        None (Baseline)     & 48.8 & 41.8 & 0.00 & 0.0\% \\
        Repeat ($\mathcal{C}_{\mathrm{rep}}$) & 57.0 & 45.1 & 88 & 4.74\% \\
        \rowcolor{gray!10}
        \textbf{Deviation ($\mathcal{C}_{\mathrm{away}}$)} & \textbf{66.0} & \textbf{49.4} & 279 & 15.0\% \\
        Hybrid (Both)       & 54.0 & 44.5 & 477 & 23.1\% \\
        Periodic ($k=5$)    & 63.0 & 46.5 & 113 & 10.2\% \\
        \midrule
        Human-in-the-Loop   & 74.5 & 61.2 & 105 & 9.8\% \\
        \bottomrule
        \end{tabular}
        }
    \end{minipage}
    \begin{minipage}{\linewidth}
        \centering
        \captionsetup{type=figure} 
        \includegraphics[width=\linewidth]{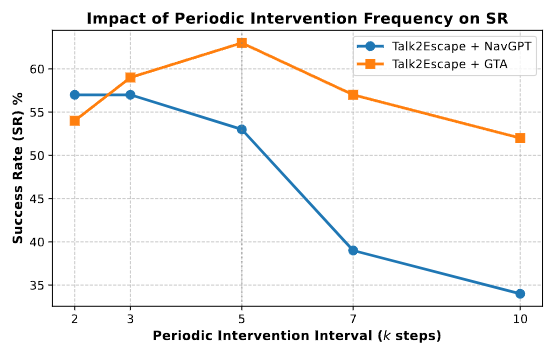} 
        \vspace{-2em}
        \caption{\textbf{Impact of Intervention Frequency on Different Agents:} Comparison of navigation performance under varying periodic intervention intervals ($k$). The evaluation contrasts the performance trends of a vanilla baseline agent (NavGPT) with a more complex agent (GTA).}
        \label{fig:periodic_interval}
        \vspace{-2em}
    \end{minipage}
    
\end{figure}

\subsubsection{Model-Agnosticism in High-Fidelity Environments}

To comprehensively address RQ2 and verify the generalizability of our framework, we conduct additional ablations on the fine-grained instruction task of the high-fidelity \textbf{VLNVerse}~\cite{lin2025vlnverse} benchmark. This environment provides superior physical simulation and richer visual rendering, serving as a rigorous testbed for MLLM perception. We evaluate the native open-loop baselines provided by this benchmark, specifically \textbf{NavGPT}~\cite{zhou2024navgpt} and \textbf{MapGPT}~\cite{chen2024mapgpt}. To determine whether passive internal reasoning can substitute for active external dialogue, we test their Chain-of-Thought (CoT) variants alongside our integrated \textit{Talk2Escape} (\textbf{T2E}) module. As presented in Tab.~\ref{tab:vlnverse_abl}, the integration of T2E yields staggering improvements that completely eclipse the marginal gains of passive CoT reasoning. For instance, while CoT improves NavGPT's Success Rate (SR) by only 9.40\%, our proactive dialogue intervention skyrockets the SR from 19.30\% to an impressive 84.60\%, with SPL surging from 8.13 to 35.20. MapGPT exhibits a similarly massive +41.49\% absolute SR gain when equipped with T2E. This conclusively demonstrates the model-agnostic efficacy of our framework across structurally distinct base agents.

Interestingly, while MapGPT's explicit 3D coordinate tracking provides a slight advantage in the unassisted open-loop setting (25.53\% SR compared to NavGPT's 19.30\%), it scales significantly worse when integrated with the T2E module (67.02\% SR versus NavGPT's 84.60\%). We attribute this counter-intuitive result to a profound modality mismatch. Overwhelming the MLLM with dense, absolute 3D numerical coordinates conflicts with the relative, egocentric semantics of the Oracle's dialogue hints. MapGPT struggles to reconcile these relative instructions with its absolute coordinate history, which diminishes its recovery rate. This reveals a crucial architectural insight: for interactive, dialogue-driven navigation, a clean and action-centric history aligns far more with natural language corrections than over-engineered geometric prompts.


\subsubsection{Effectiveness of Intervention Triggers}
\label{subsubsec:trigger_ablation}
To answer RQ3, we isolate the \textit{Talk2Escape} strategy and ablate the dialogue trigger conditions. Tab.~\ref{tab:ablation_triggers} illustrates the critical trade-off between navigation success and the frequency of guidance.

First, the \textit{Deviation} ($\mathcal{C}_{\mathrm{away}}$) strategy emerges as the most effective autonomous trigger (SR 66.0\%, SPL 49.4). It validates our design that kinematic-aware, targeted interventions provide higher-quality context than blind, scheduled polling. Crucially, the \textit{Hybrid} strategy reveals the detrimental effect of over-querying. Despite combining both \textit{Repeat} and \textit{Deviation} mechanisms, it triggers too frequently (23.1\% rate) and suffers a severe performance drop (SR 54.0\%). This confirms that excessive querying fundamentally disrupts the agent's contextual memory and leads to ``intervention fatigue.''

This phenomenon is further elucidated in Fig.~\ref{fig:periodic_interval}, which explicitly maps intervention frequency ($k$) against the capabilities of different base agents. For a vanilla baseline like NavGPT, performance degradation is roughly linear with respect to sparsity. Lacking a robust internal representation, this model relies heavily on constant external guidance. Therefore, sparser interventions at larger $k$ intervals strictly diminish its success. In stark contrast, GTA, a fundamentally stronger agent equipped with robust internal planning, exhibits a distinct inverted U-shape. For GTA, overly frequent polling ($k < 5$) severely disrupts its internal contextual memory and counterfactual reasoning loop, creating cognitive interference rather than assistance. Conversely, overly sparse polling deprives it of critical rescue. Consequently, a moderate interval ($k=5$) yields the optimal balance. This divergent behavior conclusively reinforces why the sparse and highly targeted nature of the \textit{Deviation} trigger is the superior paradigm. It avoids disrupting the autonomy of capable agents while still providing critical safety nets when physical failure is imminent.

Notably, while the synchronous \textit{Human-in-the-Loop} setup provides ground-truth subjective guidance, it suffers from the perceptual discontinuity inherent to discrete simulator rendering, leading to delayed or sub-optimal interventions. This confirms our hypothesis: automated, kinematic-aware triggers offer a more robust and less fatiguing safety net than relying entirely on discrete human visual polling.

\subsection{Real-World Robot Deployment}
\label{subsec:real_world}
\begin{figure}[t]
\vspace{0.5em}
  \centering
  \setlength{\abovecaptionskip}{2pt}
  \setlength{\belowcaptionskip}{3pt}
  \includegraphics[width=0.98\linewidth]{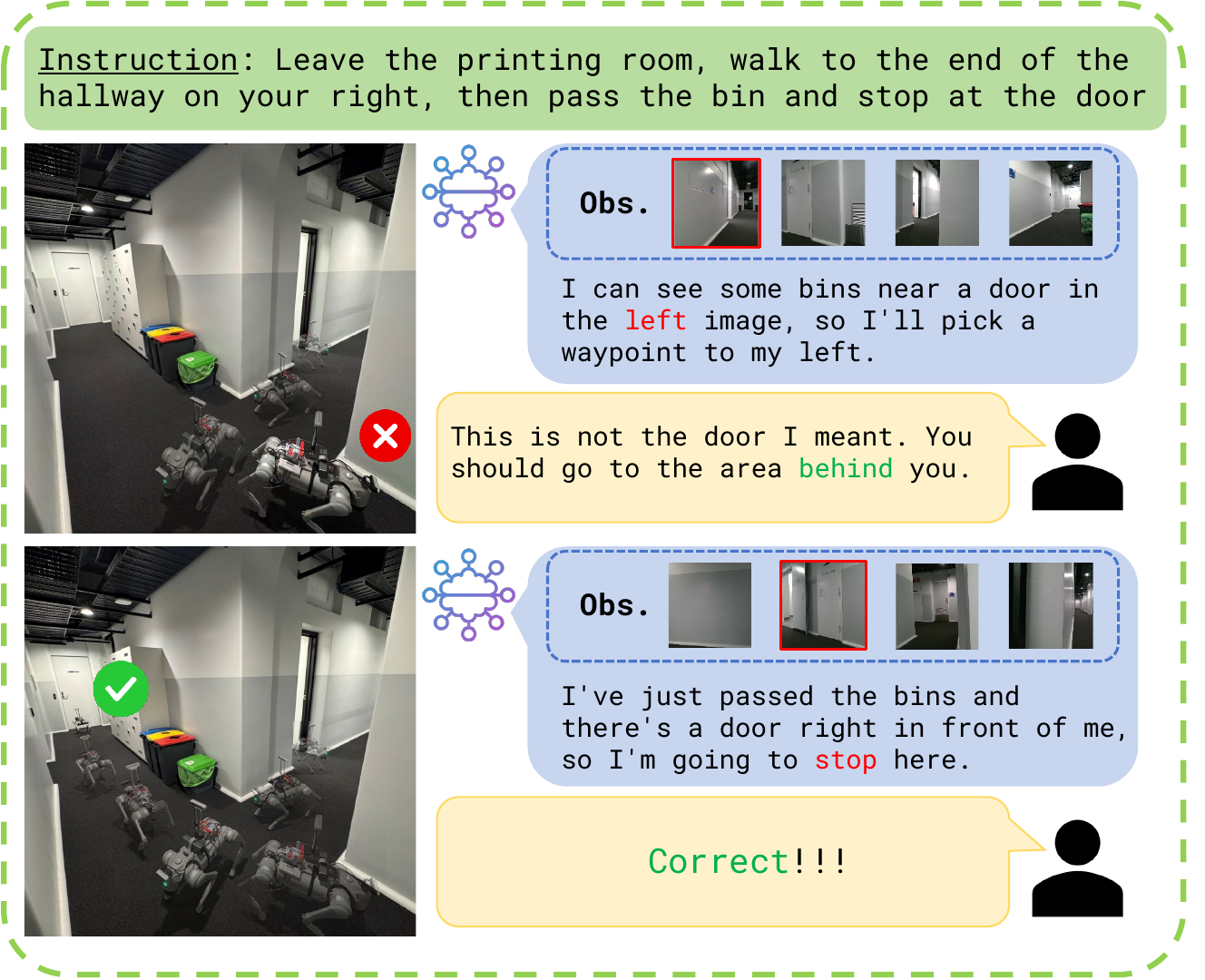}
  \caption{\textbf{Real-World Execution of the \textit{Talk2Escape} Paradigm.} \textbf{(Top)} The agent misinterprets the instruction and navigates toward an incorrect door (red cross), prompting a human guide to provide a grounded corrective hint via dialogue. \textbf{(Bottom)} Empowered by this feedback, the agent successfully corrects its route, reaches the intended target (green checkmark), and receives final confirmation.}
  \label{fig:demo}
  \vspace{-2em}
\end{figure}

To address RQ4, we deploy the \textit{Talk2Escape} framework on a physical Unitree Go2 quadruped robot across diverse, unmapped real-world scenarios such as Fig.~\ref{fig:demo}. Unlike simulated environments, physical deployment introduces severe embodied challenges, including asynchronous control latency, dynamic obstacles, reflective surface noise, and inevitable odometry drift. During our physical trials, we employ a strict human-in-the-loop (HITL) monitoring protocol for trigger activation. A human supervisor continuously monitors the real-time visual and odometry streams, but crucially, only halts the robot and injects a corrective dialogue when the Go2 exhibits severe kinematic slippage or trajectory deviation. This protocol strictly mirrors the \textit{Deviation} ($\mathcal{C}_{\mathrm{away}}$) trigger condition validated in simulation. Upon receiving the text, the onboard MLLM successfully re-plans the heading, effectively rescuing the agent from terminal physical failures and substantially increasing overall navigation success.

Consequently, even when a human-in-the-loop receives these fragmented observations, they occasionally struggle to fully mentally reconstruct the robot's exact global orientation or local obstacle topology. This partial observability can lead to suboptimal or ambiguous corrective prompts, degrading the recovery efficiency. This critical finding indicates that relying solely on instantaneous static frames is insufficient for complex physical interventions. Future work must explore integrating continuous spatio-temporal video buffers or multi-view historical memory into the dialogue state, ensuring the guiding Oracle (or human operator) possesses a temporally coherent understanding of the robot's physical predicament. 
\section{Conclusion}
\label{sec:conclusion}
In this work, we introduced \textit{Talk2Escape}, a dialogue-driven VLN framework that reframes multi-turn interaction from passive instruction following to proactive error recovery. Extensive evaluations show that \textit{Talk2Escape} revitalizes diverse base agents and outperforms supervised state-of-the-art models entirely zero-shot. Furthermore, our ablation studies highlight the efficacy of strategic sparsity: targeted, deviation-based interventions prevent redundant querying and yield superior recovery rates compared to rigid schedules. Crucially, physical deployment on a quadruped robot validates its sim-to-real transferability, proving that grounded dialogue effectively rescues agents from severe real-world navigation failures. Future work will address the ``asymmetric context'' bottleneck via continuous spatio-temporal memory, and explore richer communication modalities for highly dynamic environments.

\bibliographystyle{IEEEtran}
\bibliography{IEEEfull}
\end{document}